\documentclass[acmtog]{acmart}

\acmSubmissionID{0}

\usepackage{mathtools} 
\usepackage{siunitx}    %
\usepackage{multirow}
\usepackage{subcaption}

\usepackage[ruled]{algorithm2e} %

\SetAlFnt{\small}
\SetAlCapFnt{\small}
\SetAlCapNameFnt{\small}
\SetAlCapHSkip{0pt}

\acmJournal{TOG}

\copyrightyear{2026}
\acmYear{2026}
\setcopyright{cc}
\setcctype{by}
\acmConference[SIGGRAPH Conference Papers '26]{Special Interest Group on Computer Graphics and Interactive Techniques Conference Conference Papers}{July 19--23, 2026}{Los Angeles, CA, USA}
\acmBooktitle{Special Interest Group on Computer Graphics and Interactive Techniques Conference Conference Papers (SIGGRAPH Conference Papers '26), July 19--23, 2026, Los Angeles, CA, USA}
\acmDOI{10.1145/3799902.3811049}
\acmISBN{979-8-4007-2554-8/2026/07}

\newif\ifdraft
\draftfalse

\ifdraft
    \newcommand{\sagi}[1]{\textcolor{blue}{[\textbf{Sagi:} #1]}}

    \newcommand{\gmc}[1]{\textcolor{orange}{[\textbf{gal:} #1]}}

    \newcommand{\dcc}[1]{\textcolor{red}{[\textbf{DC:} #1]}}

    \newcommand{\rgc}[1]{\textcolor{magenta}{[\textbf{RG:} #1]}}

    \newcommand{\revision}[1]{{\color{teal} #1}}

\else
    \newcommand{\sagi}[1]{}

    \newcommand{\gmc}[1]{}

    \newcommand{\dcc}[1]{}

    \newcommand{\rgc}[1]{}

    \newcommand{\revision}[1]{{\color{black} #1}}

\fi

\title{Video Analysis and Generation via a Semantic Progress Function}

\author{Gal Metzer}
\authornote{Both authors contributed equally to the paper}
\affiliation{%
 \institution{Tel Aviv University}
 \country{Israel}
}
\email{gal.metzer@gmail.com}

\author{Sagi Polaczek}
\authornotemark[1]
\affiliation{%
 \institution{Tel Aviv University}
 \country{Israel}
}
\email{sagi.polaczek@gmail.com}

\author{Ali Mahdavi-Amiri}
\affiliation{%
 \institution{Simon Fraser University}
 \country{Canada}
}
\email{a.mahdavi.amiri@gmail.com}

\author{Raja Giryes}
\affiliation{%
 \institution{Tel Aviv University}
 \country{Israel}
}
\email{raja@tauex.tau.ac.il}

\author{Daniel Cohen-Or}
\affiliation{%
 \institution{Tel Aviv University}
 \country{Israel}
}
\email{cohenor@gmail.com}

\begin{document}

\begin{abstract}
Transformations produced by image and video generation models often evolve in a highly non-linear manner: long stretches where the content barely changes are followed by sudden, abrupt semantic jumps. To analyze and correct this behavior, we introduce a Semantic Progress Function, a one-dimensional representation that captures how the meaning of a given sequence evolves over time. For each frame, we compute distances between semantic embeddings and fit a smooth curve that reflects the cumulative semantic shift across the sequence. Departures of this curve from a straight line reveal uneven semantic pacing. Building on this insight, we propose a semantic linearization procedure that reparameterizes (or retimes) the sequence so that semantic change unfolds at a constant rate, yielding smoother and more coherent transitions.
Beyond linearization, our framework provides a model-agnostic foundation for identifying temporal irregularities, comparing semantic pacing across different generators, and steering both generated and real-world video sequences toward arbitrary target pacing.
\end{abstract}


\begin{teaserfigure}
  \includegraphics[width=\textwidth]{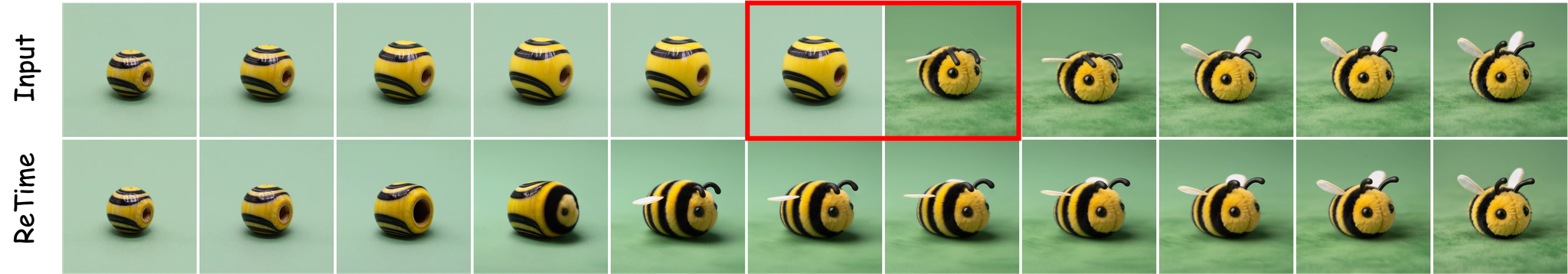}
  \caption{\textbf{From Bead to Bee.} An input generated by a video model (top) experiences an abrupt change from a bead to a bee (marked frames). Our method regenerates the video to enforce an approximately linear progression, producing smooth, evenly paced transitions (bottom) compared to the original (top).}
  \label{fig:teaser}
\end{teaserfigure}

\maketitle

\section{Introduction}

Generative models increasingly produce image and video sequences intended to depict gradual transformations-such as morphs, edits, style transitions, or object evolution. Yet these sequences often change in meaning unevenly: long stretches with almost no semantic variation are followed by abrupt jumps (see the red-marked frames in Figure~\ref{fig:teaser}) where the transformation suddenly “catches up.” This non-linear semantic evolution undermines perceptual coherence, reduces controllability, and complicates downstream editing.
\revision{This setting is not merely academic: first-last frame generation is actively used for artistic VFX, cinematic transitions, looping videos, and product reveals, and is featured in recent commercial tools.}

While prior work has addressed temporal smoothness or latent-space interpolation, these approaches do not quantify how semantic content itself evolves along a given sequence.
In particular, there is no principled measure that captures the rate of semantic change, identifies where abrupt shifts occur, or allows comparing the semantic pacing of sequences produced by different models.
A tool of this kind would provide both diagnostic insight and a foundation for improving generative stability.

In this work, we introduce a novel conceptual tool, which we call the Semantic Progress Function (SPF), for characterizing how meaning evolves over time in a video sequence.
The SPF is a one-dimensional function that represents the cumulative semantic state of a sequence, with its slope reflecting the instantaneous rate of semantic change.
It is constructed by computing semantic distances between frames and fitting a smooth curve that best reflects their ordering in meaning.
Departures of this curve from a straight line directly reveal uneven semantic pacing, pinpointing where and by how much a sequence deviates from linear semantic evolution.
As such, the SPF provides an interpretable, model-agnostic representation that makes semantic progression explicit and measurable, enabling principled analysis of generative transformations.

Building on this analysis, we propose semantic linearization, a method that reparameterizes the sequence so that semantic progress increases at a constant rate.
This correction produces sequences in which the transformation unfolds more smoothly and predictably, avoiding sudden accelerations and improving perceptual consistency.
Because the remedy is derived directly from the semantic progress structure, it is principled and requires no model fine-tuning.

Using the SPF as our analysis tool, we study the semantic evolution of a wide range of video sequences, including both synthetically generated and real-world footage.
Leveraging this analysis, we apply several retiming strategies derived directly from the SPF, demonstrating how measured semantic progress can be used to systematically correct uneven evolution and to enforce desired pacing behaviors.
We validate the proposed framework through extensive experiments and comparisons, highlighting the robustness, generality, and practical impact of the SPF as a foundation for analyzing and improving semantic temporal behavior in video sequences.

\section{Related Work}

A wide range of prior work has explored how to generate smooth transitions between visual states, spanning image morphing, video generation, and temporally controlled synthesis. Despite differences in representation and modeling assumptions, these methods share a common goal: producing perceptually coherent and semantically plausible intermediate content between given endpoints. Our work relates to these efforts, but focuses on an aspect that has received comparatively little attention—how semantic change unfolds over time within a sequence, and how uneven semantic pacing can be analyzed and corrected independently of the underlying generative model. The following sections review related work from this perspective, from classical morphing techniques to modern diffusion-based video models and temporal control mechanisms.

\noindent \textbf{Image Morphing} techniques have evolved from geometric deformations to sophisticated generative approaches. \emph{Classical Morphing} techniques relied heavily on warping and cross-dissolving. Feature-Based Image Metamorphosis~\cite{beier1992feature} utilized line segments to define correspondence fields, though they often required laborious manual annotation. Moving Least Squares (MLS)~\cite{schaefer2006image} eventually became a gold standard for deformation, producing smooth mappings from sparse control points. Subsequent comparative studies have analyzed the trade-offs between triangulation-based and feature-based methods~\cite{bhatt2011comparative}, highlighting the difficulties in handling complex geometric changes. To address artifacts such as ghosting, later works explored patch-based synthesis and regenerative morphing to automate transitions~\cite{shechtman2010regenerative, liao2014automating}.

The advent of \emph{Deep Learning} shifted the paradigm from geometric warping to latent space interpolation. Generative Adversarial Networks (GANs), such as StyleGAN, demonstrated that traversing the latent space of a generator could yield smooth image sequences \cite{karras2019style}. Works focusing on perceptual constraints and Spatial Transformer Network (STN) alignment further refined these transitions~\cite{fish2020image}. A significant leap was achieved with Alias-Free GANs, which offered rotation and translation invariance, making interpolations more structurally consistent \cite{karras2021alias}. To apply these capabilities to real images, inversion techniques such as deep generative priors and pixel2style2pixel (pSp)~\cite{richardson2021encoding, tov2021designing, alaluf2021restyle} encoders were developed to project images into the GAN latent space for manipulation.
Most recently, \emph{Diffusion Models} have become the state-of-the-art for morphing. Unlike GANs, diffusion models offer more stable training and higher semantic fidelity. While \cite{wang2023interpolating} demonstrated that diffusion models could interpolate by navigating noise space, contemporary methods have significantly advanced this capability. DiffMorpher~\cite{zhang2023diffmorpher} and FreeMorph~\cite{cao2025freemorph} employ techniques such as LoRA-based fine-tuning and attention control mechanisms to ensure smooth semantic transitions without the need for extensive training on specific concept pairs.

Unlike static morphing, video generative models hallucinate realistic motion and dynamics. Our approach leverages this to synthesize complex motion (e.g., fluids, lighting) that geometric warping cannot capture.

\noindent \textbf{Video Diffusion Models} (VDMs) extend image synthesis to the temporal dimension, introducing challenges in maintaining consistency across frames~\cite{yan2023temporally, kim2024stream}. 
Foundational works have focused on generating coherent motion from textual or image prompts. 
A critical capability for morphing applications within VDMs is "first-last frame" conditioning, as seen in models like Wan~\cite{wan2025} and LTX-Video~\cite{HaCohen2024LTXVideo}.
This approach constrains generation between specific start and end images, creating a bridge between static image morphing and dynamic video generation. This allows the model to hallucinate plausible motion and semantic transitions between two distinct endpoints.

\noindent \textbf{Temporal Controllability in Video Generation}
is an active research area that explores achieving precise control over the timing and speed of generated video content.
Recent approaches have introduced mechanisms to modulate temporal attention.
Techniques utilizing Rotary Positional Embeddings (RoPE)~\cite{su2023roformerenhancedtransformerrotary} modulations allow for the scaling of temporal dynamics, effectively stretching or compressing motion without retraining \cite{wei2025videorope, gokmen2025ropecraft, zhao2025riflexfreelunchlength}.
\revision{LoViC~\cite{jiang2025lovic} applies RoPE modulation for context compression in long video generation; in contrast, our work warps temporal positions based on measured semantic content via the SPF, with per-band frequency control to correct pacing rather than extend context length.}
Furthermore, methods like TempoControl~\cite{schiber2025tempocontrol} introduce temporal attention guidance, which explicitly manipulates cross-attention maps to align specific video frames with distinct parts of the text prompt. 

However, TempoControl necessitates manual user intervention in the form of spatial masks to define exactly where each text token should influence the video generation.
In contrast, our approach enables the transformation of both generated and "in-the-wild" videos into a constant pace without requiring any manual user annotation.
Furthermore, we introduce a novel pace-measuring metric that allows for the objective quantification of temporal linearity, a capability not addressed by existing guidance-based methods.

\section{Semantic Progress Function}
\label{sec:method}

\begin{figure*}[t]
    \centering
    \includegraphics[width=\textwidth]{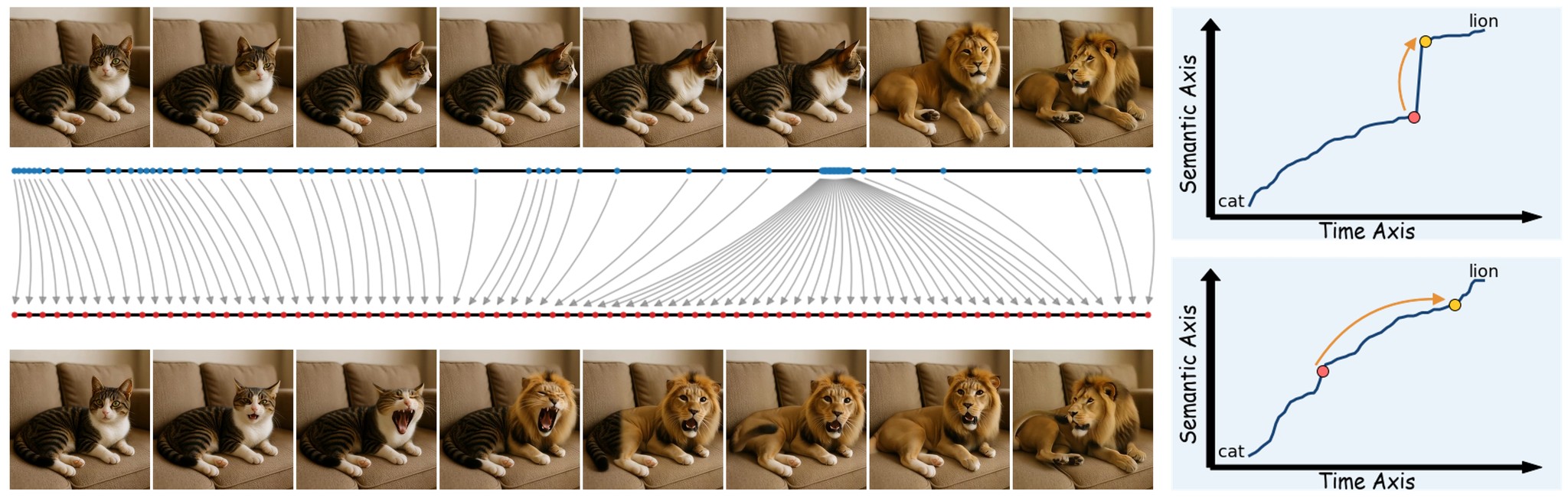}
    \caption{\textbf{ReTime Overview.} 
    The top row shows an input sequence with an abrupt semantic shift, reflected by the discontinuity in the \textit{semantic progress function} (top right). 
    The center diagram visualizes the retiming as performed on the RoPE embeddings, where input time embeddings (blue) are warped in order to \textit{linearize} the output timestamps (red). 
    The bottom row demonstrates the retimed result, achieving a constant semantic pace as shown by the linearized function (bottom right).}
    \label{fig:retiming_method_main}
\end{figure*}

In this section, we formally define the Semantic Progress Function (SPF), a model-agnostic formulation of semantic evolution over time. The SPF serves as our central representation, distilling complex visual transformations into a one-dimensional trajectory that allows analyzing diverse models and domains. 
Formally, given a video consisting of $T$ frames $\{x_1, x_2, \dots, x_T\}$, we define the SPF as a scalar-valued function $S_i \in \mathbb{R}$ mapped from the frame index $i$. This concept is visualized in Figure~\ref{fig:retiming_method_main}, which displays sample video frame strips alongside their corresponding SPF trajectories on the right. 
The construction proceeds in two stages: we first compute pairwise semantic distances between frames, and then integrate these differences over time. By design, the SPF is integrated such that differences in its value approximate the semantic distances between video frames. Consequently, $S_i$ represents the cumulative semantic state of the video at frame $i$, while the slope of the function reflects the instantaneous rate of semantic change.

\subsection{Frame-Level Semantic Distance Computation}
The initial part of the SPF paradigm assumes a pairwise model (or oracle) to measure the semantic differences between image pairs. 
This design choice is driven by the abundance of pretrained semantic image embedders developed by the computer vision community~(e.g., CLIP~\cite{radford2021learningtransferablevisualmodels}, DINO~\cite{siméoni2025dinov3}).
Such models embed images into a latent space where dot product measures semantic similarity, making them highly suitable for measuring similarity between image pairs.
For our ReTime technique (Section~\ref{sec:retime}), we choose SigLIP~\cite{zhai2023sigmoid} due to its strong performance for our downstream applications.

In more detail, each video frame $x_i$ is mapped to a semantic embedding $z_i \in \mathbb{R}^d$ using SigLIP.
Semantic distances between frames are computed using an angular metric in the embedding space.
Specifically, embeddings are $\ell_2$-normalized and the distance between frames $i$ and $j$ is defined as:

\begin{equation}
\label{eq:angular_distance}
d_{ij} = \arccos \left( z_i^\top z_j \right),
\end{equation}

While the formulation permits the use of semantic distances between all frame pairs in the video, this is not required in practice.
For computational efficiency and to emphasize local temporal structure, we restrict the set of pairs $\mathcal{P}$ to frames whose temporal distance satisfies $|i - j| \leq 30$. 
The resulting set of distances $\{d_{ij}\}$ is used as supervision for fitting the SPF.

\subsection{Fitting the Semantic Progress Function (SPF)}

We seek to estimate the SPF as a vector $S \in \mathbb{R}^T$, where $S_i$ denotes the value at frame $i$ and $T$ is the number of frames.
As previously stated, $S_i$ ought to be constructed such that its pairwise temporal differences approximate the semantic distances between video frames:
\begin{equation}
\label{eq:seek_potential}
S_i - S_j \approx d_{ij}, \quad \forall (i,j) \in \mathcal{P} : i > j
\end{equation}
where $\mathcal{P}$ denotes a set of frame pairs (e.g., all pairs or a selected subset). To express these relations compactly, let $M = |\mathcal{P}|$ be the number of pairs, let $b \in \mathbb{R}^M$ collect the distances $d_{i_k j_k}$, and construct $A \in \mathbb{R}^{M \times T}$ whose $k$-th row corresponding to pair $(i_k,j_k)$ has a $+1$ at index $i_k$, a $-1$ at index $j_k$, and zeros elsewhere. With this notation we may rewrite Eq.~\eqref{eq:seek_potential} via the following linear system:
\begin{equation}
A S \approx b.
\end{equation}

We estimate $S$ via a regularized, weighted least-squares objective:
\begin{equation}
\min_{S \in \mathbb{R}^T} \; (A S - b)^\top W (A S - b) \;+\; \lambda \, S^\top S,
\label{eq:wls-matrix}
\end{equation}
where $W \in \mathbb{R}^{M \times M}$ is a diagonal matrix with entries $W_{kk} = w_{i_k j_k} \ge 0$ weighting the constraint for the pair $(i_k,j_k)$, and $\lambda > 0$ controls the regularization strength. 
We design weights to favor temporally local constraints using a Gaussian function of temporal distance:
\begin{equation}
w_{ij} = \exp\!\left( -\frac{(i - j)^2}{2\sigma^2} \right),
\end{equation}
where $\sigma$ sets the temporal scale over which constraints are emphasized.

For $\lambda > 0$, objective~\eqref{eq:wls-matrix} is strictly convex and has the unique closed-form solution:
\begin{equation}
\hat{S} = (A^\top W A + \lambda I)^{-1} A^\top W b.
\end{equation}
An example of such SPFs is visualized on the right side of Figure~\ref{fig:retiming_method_main}. The top graph depicts the SPF of the raw input video. Notably, the rate of change in $S$ increases sharply where the cat abruptly transforms into a lion (indicated by the orange arrow), reflecting the semantic discontinuity of the video. The bottom graph illustrates the SPF after retiming, where the semantic progression appears significantly steadier as expected.

\section{Video Linearization via ReTime}
\label{sec:retime}

While the Semantic Progress Function (SPF) provides a diagnostic view of how semantic change unfolds over time, it also enables direct intervention.
In this section, we leverage the SPF to reparameterize time so that semantic change progresses at a constant rate, a process we refer to as semantic \textit{linearization}.
This correction redistributes temporal capacity according to measured semantic change, producing smoother and more predictable transformations without retraining or manual annotation.
Below, we provide details on how the SPF is utilized to retime generated and existing videos.

\subsection{Retiming of Generated Videos}
\label{subsec:controlled_regen}
\begin{figure}[t]
    \centering
    \includegraphics[width=0.9\columnwidth]{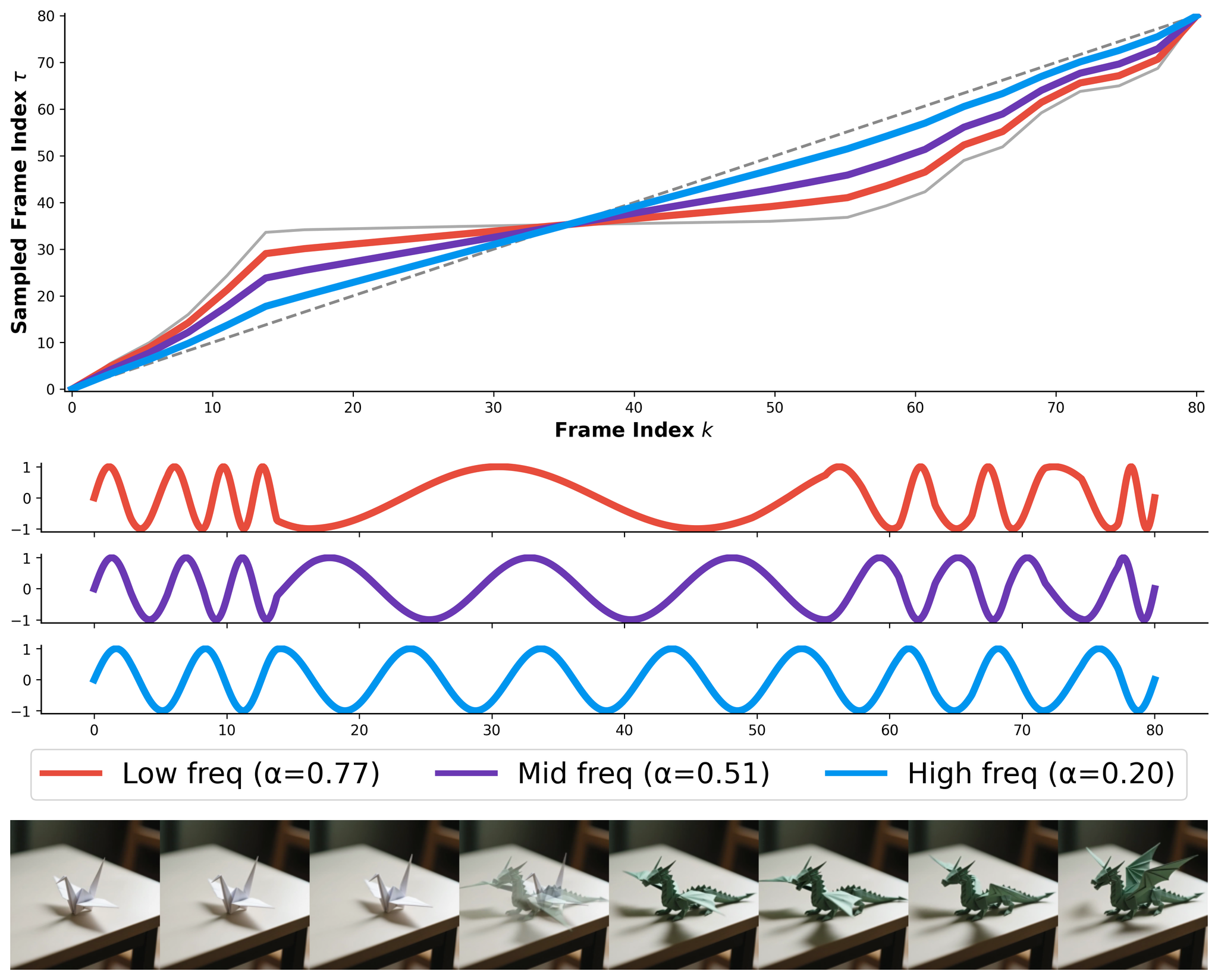}
\caption{\textbf{Frequency-Aware Retiming.} 
\textbf{(Top)} Retiming schedule maps the \textbf{Frame Index $k$} to the \textbf{Sampled Frame Index $\tau$}. The \textbf{target schedule} (solid grey) creates a plateau effectively slowing down the fast semantic jump.
\textbf{Low-frequency bands} (red, $\alpha=0.77$) strictly track this schedule to correct global pacing, while \textbf{High-frequency bands} (blue, $\alpha=0.20$) remain nearly linear to preserve local motion smoothness. 
\textbf{(Middle)} Waveforms illustrate the mechanism: low frequencies spatially warped (stretched) to dilate time.
\textbf{(Bottom)} The \textbf{Input Video} sequence being analyzed.}

\label{fig:rope_retime}
\end{figure}

Given a video diffusion model, we first generate a transformation sequence and analyze its Semantic Progress Function.
In many cases, the resulting semantic change is not temporally uniform, with long intervals that exhibit little perceptual change followed by abrupt transitions.
We therefore regenerate the sequence with an explicit retiming mechanism that warps the model’s temporal positional encodings according to the measured progress curve, allocating more temporal capacity to semantically dense regions and less to stable ones.
This intervention is applied at inference time, requires no retraining, and avoids post-hoc interpolation.
Figure~\ref{fig:retiming_method_main} illustrates the method visually, demonstrating time-warping alongside the before-and-after analysis.

\paragraph{Temporal Position Warping.}
The SPF $S$ maps frame indices to normalized cumulative progress (via min-max scaling to $[0,1]$), $S: \{1,\ldots,T\} \to [0,1]$. For uniform semantic velocity, output frame $k$ should exhibit progress $k/(T{-}1)$. We achieve this by computing warped temporal positions via inversion:
\begin{equation}
\label{eq:tau}
\tau_k = S^{-1}\!\left(\frac{k}{T{-}1}\right),
\end{equation}

where $S^{-1}$ denotes piecewise-linear interpolation over the discrete samples. This stretches time in regions of rapid semantic change and compresses stable regions, redistributing temporal capacity according to perceptual importance.
Figure~\ref{fig:retiming_method_main} (center) visualizes this process, showing how input time embeddings (blue) are warped to achieve linear output pacing (red).

\paragraph{RoPE Integration.}
Modern video diffusion transformers such as Wan~\cite{wan2025} employ Rotary Position Embeddings~\cite{su2023roformerenhancedtransformerrotary} along the temporal axis. RoPE encodes position $p$ by applying frequency-dependent rotations to query and key vectors in attention:
\begin{equation}
\mathbf{q}_p = R_\theta(p)\,\mathbf{q}, \quad \mathbf{k}_p = R_\theta(p)\,\mathbf{k},
\end{equation}
where $R_\theta(p)$ rotates by angle $\theta \cdot p$ at each frequency $\theta$. Substituting the warped positions $\tau_k$ for linear indices causes the model to perceive non-uniform temporal spacing aligned with the semantic progress structure.

\begin{figure}[t]
    \centering
    \includegraphics[width=0.9\columnwidth]{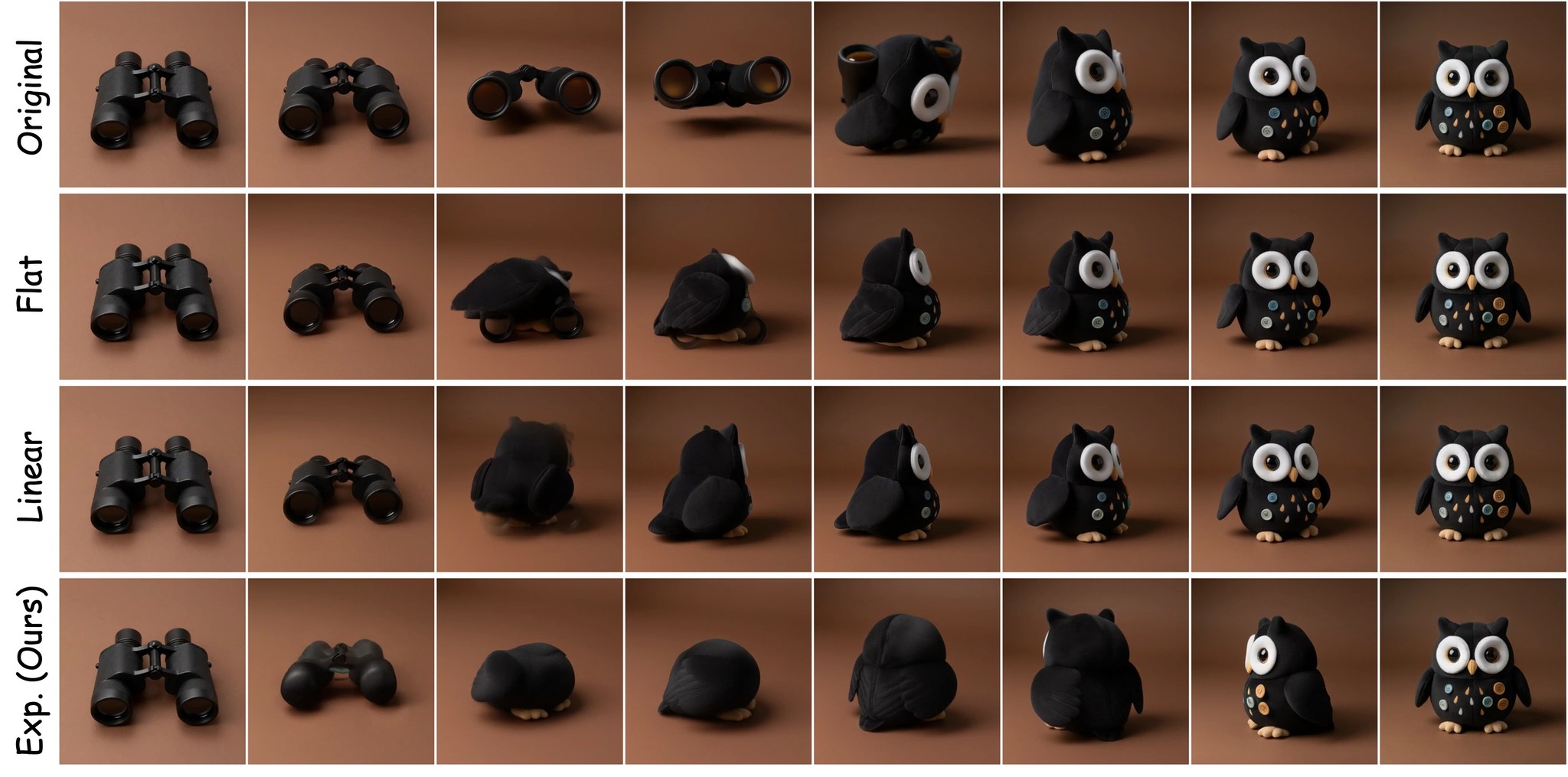}
\caption{\textbf{RoPE Frequency Schedule Ablation.} Without retiming (top), the transformation is abrupt and uneven. A flat schedule     
accelerates the transition unnaturally, while a linear schedule produces blurry intermediates. Our exponential decay schedule   
(bottom) yields a smooth, gradual transformation with coherent intermediate states.}

\label{fig:ablation_rope_freq}
\end{figure}

\paragraph{Frequency-Aware Warping.}
Temporal RoPE represents position using $B$ frequency bands, where low frequencies control long-range structure and high frequencies capture short-range dynamics.
A naive retiming would warp all bands identically. We find (Figure~\ref{fig:ablation_rope_freq}) that this can destabilize generation: warping high frequencies induces local jitter, whereas insufficient warping of low frequencies fails to correct global pacing.
We therefore introduce \emph{frequency-aware warping} by blending between the original index and the warped position with a band-dependent strength $\alpha_b \in [0,1]$:
\begin{equation}
p^{(b)}_t = (1-\alpha_b)\,t + \alpha_b\,\tau_t,
\end{equation}
where $t\in\{1,\ldots,T\}$ indexes the output frame and $\tau_t$ is defined in Eq.~\eqref{eq:tau}.
We set $\alpha_b$ to decay exponentially from low to high frequencies,
\begin{equation}
\alpha_b = \alpha_{\mathrm{high}} + (\alpha_{\mathrm{low}}-\alpha_{\mathrm{high}})\,e^{-\kappa b/(B-1)},
\label{eq:exp_schedule}
\end{equation}
so low-frequency bands receive stronger warping while high-frequency bands remain closer to linear time. 
Several alternative warping heuristics are compared in Figure~\ref{fig:ablation_rope_freq}. The approach described by Equation~\ref{eq:exp_schedule}, located in the last line, typically produces the most accurate outcomes.

\paragraph{Timestep-Dependent Modulation.}
The diffusion denoising process transitions from coarse structure (high noise) to fine details (low noise). We modulate warping strength across this trajectory via a decay multiplier $\gamma(\tilde{t}) \in [0,1]$, yielding effective per-band strength $\alpha_b^{\mathrm{eff}}(\tilde{t}) = \alpha_b \cdot \gamma(\tilde{t})$. We employ an exponential schedule that applies stronger warping early in denoising:
\begin{equation}
\gamma(\tilde{t}) = \frac{e^{3\tilde{t}} - 1}{e^3 - 1},
\end{equation}
where $\tilde{t} \in [0,1]$ is the normalized diffusion timestep ($\tilde{t}{=}1$ at maximum noise). This concentrates semantic correction during structure formation while allowing natural detail refinement at later stages.

\begin{figure*}
    \centering
    \includegraphics[width=\linewidth]{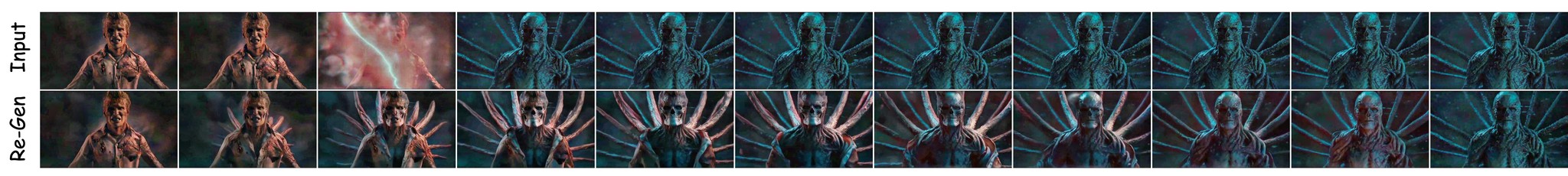}
     \caption{\textbf{Cinematic Video Linearization.}~\cite{stranger_things_vecna_2026} Two sampled frame strips near the transition: the   
     top row (original) shows a lightning-driven, abrupt change; the bottom row (linearized)              
     redistributes semantic change over time, revealing smooth intermediate stages.}                         \label{fig:timeline_normalization_vecna}
\end{figure*}

\paragraph{Iterative Refinement.}

A single warping pass may often fail to linearize semantic progress due to deviations from the target pace in the video generation process. To address this, we employ an iterative refinement scheme. Let $\tau^{(n)}$ represent the warped positions at iteration $n$. Using the current video's semantic progress function $S^{(n)}$, we compute a temporal correction:

\begin{equation}
\delta^{(n)}_k = \bigl(S^{(n)}\bigr)^{-1}\!\left(\tfrac{k}{T-1}\right) - k
\end{equation}

Positions are updated per frequency band $b$ with a step size $\alpha_b$:

\begin{equation}
\tau_k^{(n+1),(b)} = \tau_k^{(n),(b)} + \alpha_b \cdot \delta^{(n)}_k
\end{equation}

Empirically, three iterations are sufficient to achieve near-linear semantic progression.

\paragraph{Latent-Space Mapping.}
Video diffusion models operate on temporally compressed latent representations. For models using 4$\times$ temporal compression (e.g., Wan's VAE), latent step 1 corresponds to frame 1, while latent step $i$ ($i \geq 1$) corresponds to the center of frames $[4(i{-}1){+}1,\, 4i]$, i.e., frame index $4i - 1.5$. We resample the frame-level warped positions to latent resolution by interpolating at these center locations, ensuring that the temporal correction is applied in the coordinate system the model actually uses.

\paragraph{Extension to Audio-Video Models.}                            

As a direct consequence of this design, with LTX-2, the generated audio remains perfectly aligned with the retimed video output.
By anchoring cross-modal attention to linear temporal coordinates, the model generates audio that naturally
synchronizes with the semantically linearized visual content.
LTX-2 qualitative results are provided in the supplementary material.

\subsection{Retiming of Existing Videos}
\label{subsec:existing_video_linear}

When video generation is beyond our control, e.g., when the video is produced by a closed-source model or obtained from real-world sources, we propose an alternative linearization procedure. This method transforms any input video (whether captured, edited, or generated) into a temporally uniform sequence aligned with semantic progression.
The approach first segments \( S \) into piecewise linear components, then regenerates intermediate clips for each segment.
Below we detail the algorithmic stages and design choices. 

\begin{figure}[b]
    \centering
    \includegraphics[width=0.8\columnwidth]{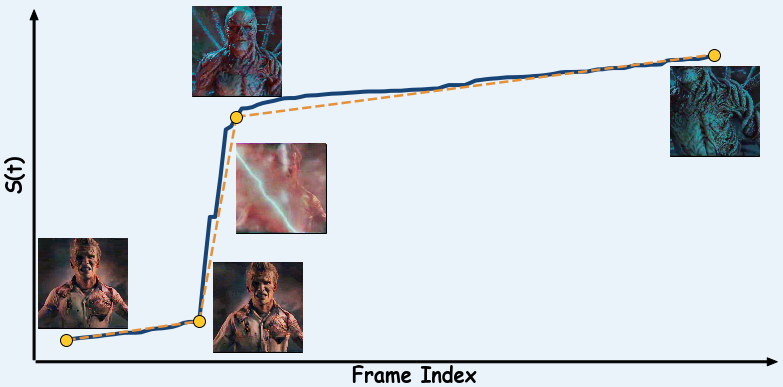}
    \caption{\textbf{SPF Segmentation.} Semantic Progress Function $S$ of the cinematic video~\cite{stranger_things_vecna_2026} shown in Figure~\ref{fig:timeline_normalization_vecna}. The steep rise marks the transition region. The dotted lines represent the least squares algorithm results, capturing the different phases of the video.}
    \label{fig:real_example_vecna}
\end{figure}

\paragraph{Timeline Segmentation.}
Given the semantic progress function $S$ over discrete frames $t \in \{1,\dots,T\}$, we apply segmented least squares to partition $S$ into $K$ contiguous, approximately linear segments $[a_k, b_k]$ such that:
\[
\begin{aligned}
&\quad 1 = a_1 \le b_1 < a_2 \le b_2 < \cdots < a_K \le b_K = T.
\end{aligned}
\]
Importantly, the segmentation is \emph{tight}: the end of each segment serves as the start of the next, ensuring full coverage of the timeline without gaps or overlaps. Segmented least squares balances fidelity to $S$ with a complexity penalty, yielding segments whose slopes reflect locally steady semantic progression. This segmentation isolates regions of near-constant semantic velocity and delineates transitions where the semantic pace differs. 
Figure~\ref{fig:real_example_vecna} shows segmented least squares results (dotted lines) on a real video from the show \textit{Stranger Things}~\cite{stranger_things_vecna_2026}.
As shown, the segments capture the different phases of the video.

\paragraph{Intermediate Clip Regeneration.}
We treat the first and last frames of each segment as semantic keyframes that are used for regeneration.
We propose two options for regeneration, one with Wan2.2 and another with LTX-2.
For LTX-2, we condition on an ordered keyframe list $\{(f_{t_i}, \hat{t}_i)\}_{i=0}^{K}$ with target frame indices defined as $\hat{t}_i \coloneqq \lfloor T S_i \rfloor$.
This formulation ensures that each frame is placed at a temporal index proportional to its cumulative semantic change.
For Wan2.2, which restricts conditioning to the \emph{first} and \emph{last} frames instead of an arbitrary sequence, we generate $K$ clips aligned with the segments.
We assign the endpoints $(f_{t_{k-1}}, f_{t_k})$ to the $k$-th clip and determine its length via $T_k \coloneqq \mathrm{round}(T\ \cdot \Delta S_k)$.
This allocation ensures that the duration of each segment is proportional to the magnitude of the semantic change between its boundary frames.
Finally, the generated clips, subject to their endpoint constraints, are concatenated to form the final video.

In general, these two paradigms allow the use of any open/closed source models, as long as they can be conditioned on either keyframes or the first-last frames, encompassing most, if not all, models available today.

\section{Experiments}

\begin{figure*}[t!]
    \centering
    \includegraphics[width=\textwidth]{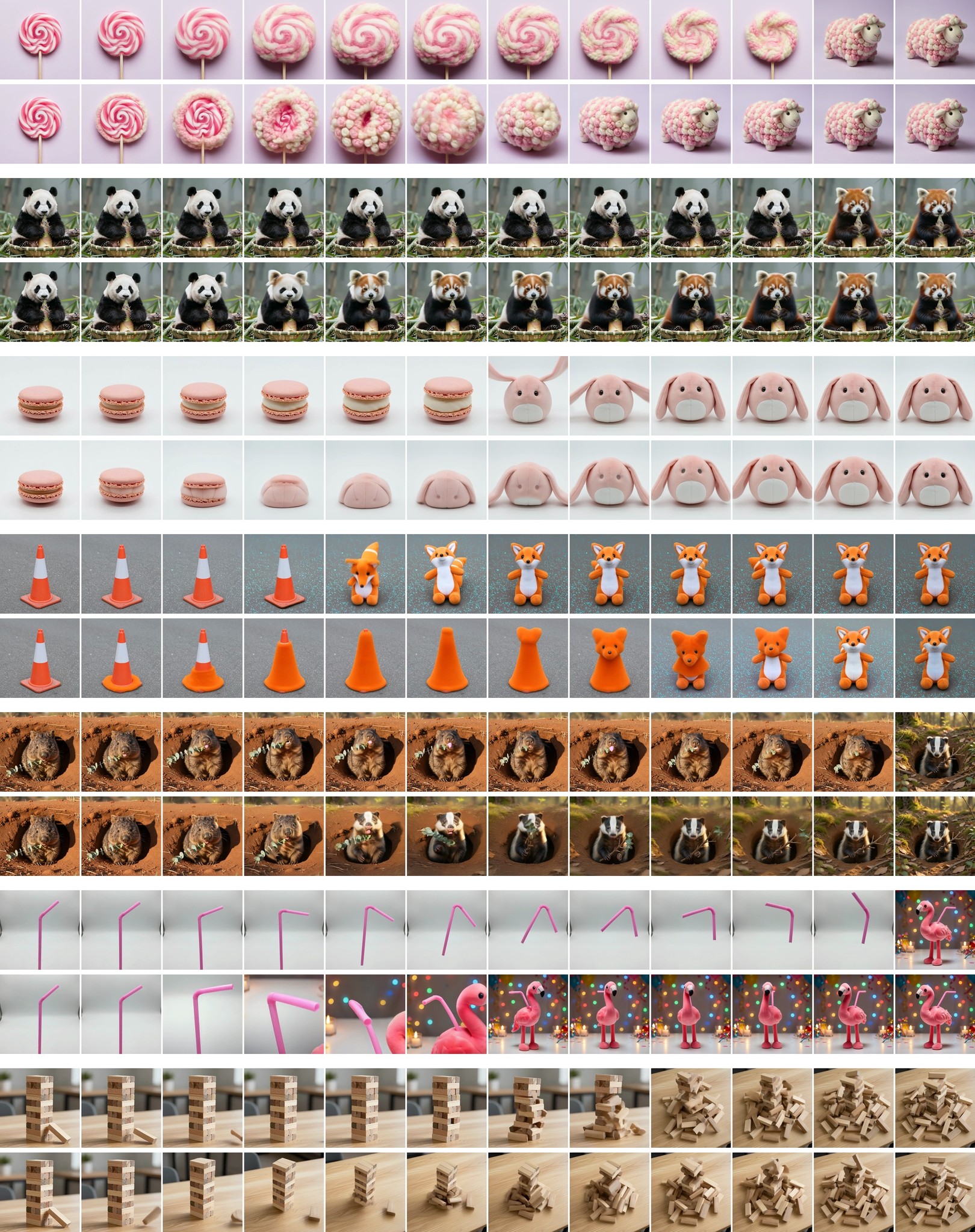}
\caption{\textbf{Qualitative Results on Wan2.2.} Selected samples of generated videos retimed using our method. The sequences showcase a variety of semantic transformations, ranging from object morphing (e.g., macarons $\to$ bunnies, cones $\to$ foxes) to physical dynamics (Jenga tower collapse). By enforcing a linear Semantic Progress Function, our method ensures these transitions unfold at a constant perceptual rate, eliminating the abrupt jumps often found in raw model outputs.}
  \label{fig:general_results_0}
\end{figure*}

We evaluate our framework through a suite of experiments designed to validate the SPF analysis and our retiming generation. 
We begin by comparing our method against baseline retiming strategies, demonstrating the advantages of time embedding intervention. We then show qualitative results on real cinematic footage.
Subsequently, we verify the SPF's accuracy using controlled synthetic experiments with known pacing profiles and analyze the SPF's sensitivity to hyperparameter choices. 
Finally, we report quantitative metrics and user study results that confirm our approach preserves visual fidelity while effectively regularizing semantic pace. 
For additional qualitative results, please refer to Figures~\ref{fig:general_results_0} and~\ref{fig:general_results_1}; further implementation details, ablations, and comparisons are available in the Supplementary Material.

\subsection{Retiming Strategy Comparison}
\begin{figure}[t]
  \centering
  \includegraphics[width=0.9\linewidth]{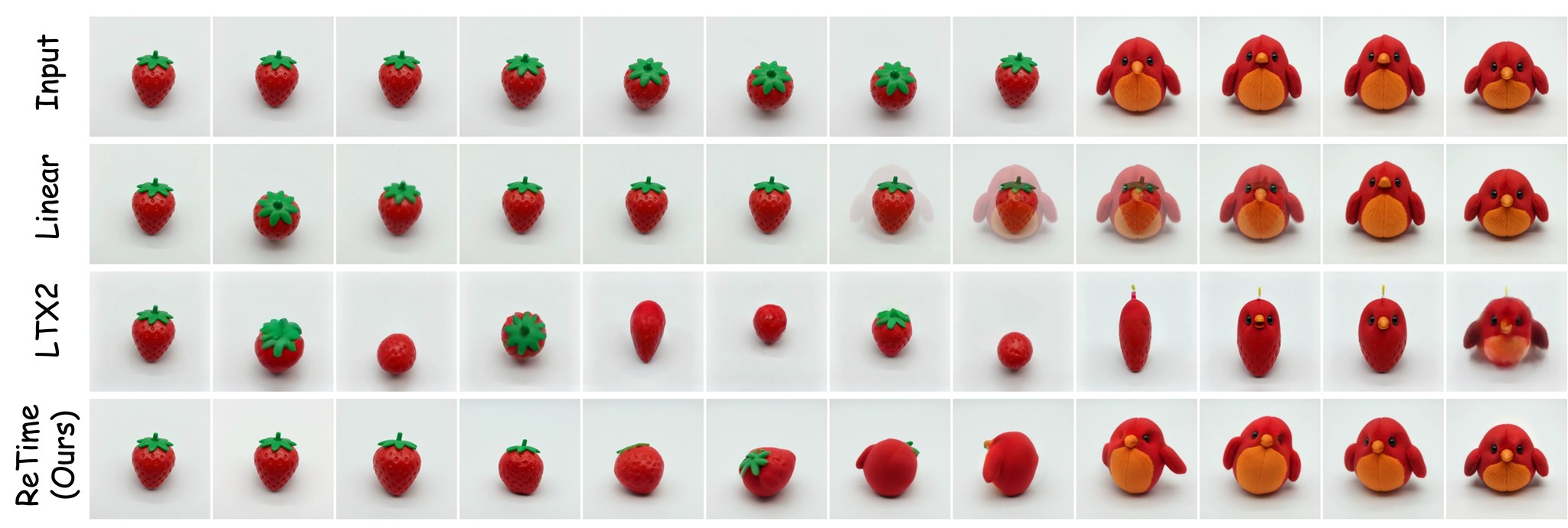}
  \caption{\textbf{Retiming Strategy Comparison.} The input sequence features a challenging semantic shift (strawberry $\rightarrow$ bird). Linear interpolation fails to resolve the transition, resulting in ghosting artifacts. Using LTX-2~\shortcite{hacohen2025ltx2} as an external keyframe-interpolator imposes a quality bottleneck, restricting results to the external model's generative limits. Ours operates directly on the input's feature representation, avoiding external quality caps and achieving superior quality.}
  \label{fig:resampling}
\end{figure}

Figure~\ref{fig:resampling} demonstrates different naive synthesis strategies for the final retiming step on a video featuring an abrupt strawberry$\rightarrow$bird transition.
Linear pixelwise interpolation (second row) fails to handle this semantic shift, resulting in ghosting.
We also compare against LTX-2~\shortcite{hacohen2025ltx2} in key-frame interpolation mode.
Relying on an external model inherently limits the output quality to that model's generative constraints.
In contrast, our method (bottom row) operates directly on the underlying model features; this preserves the intrinsic capacity of the input model without introducing external bottlenecks, yielding a significantly more coherent result.

\subsection{Real Cinematic Video}
We applied video linearization (Section~\ref{subsec:existing_video_linear}) to a transformation sequence from \textit{Stranger Things}~\shortcite{stranger_things_vecna_2026} (Figure~\ref{fig:timeline_normalization_vecna}) using Wan2.2~\shortcite{wan2025}.
As shown in Figures~\ref{fig:timeline_normalization_vecna} and ~\ref{fig:real_example_vecna}, while the original metamorphosis is obscured by an abrupt lighting cue, our method redistributes this change to create a smooth, continuous evolution, capturing the gradual growth of background elements and the steady transition from human to monster.
Additional results, including examples generated with LTX-2, are provided in the supplementary material.

\subsection{Non-Linear Retiming}

\begin{figure}[t]
  \centering
  \includegraphics[width=\linewidth]{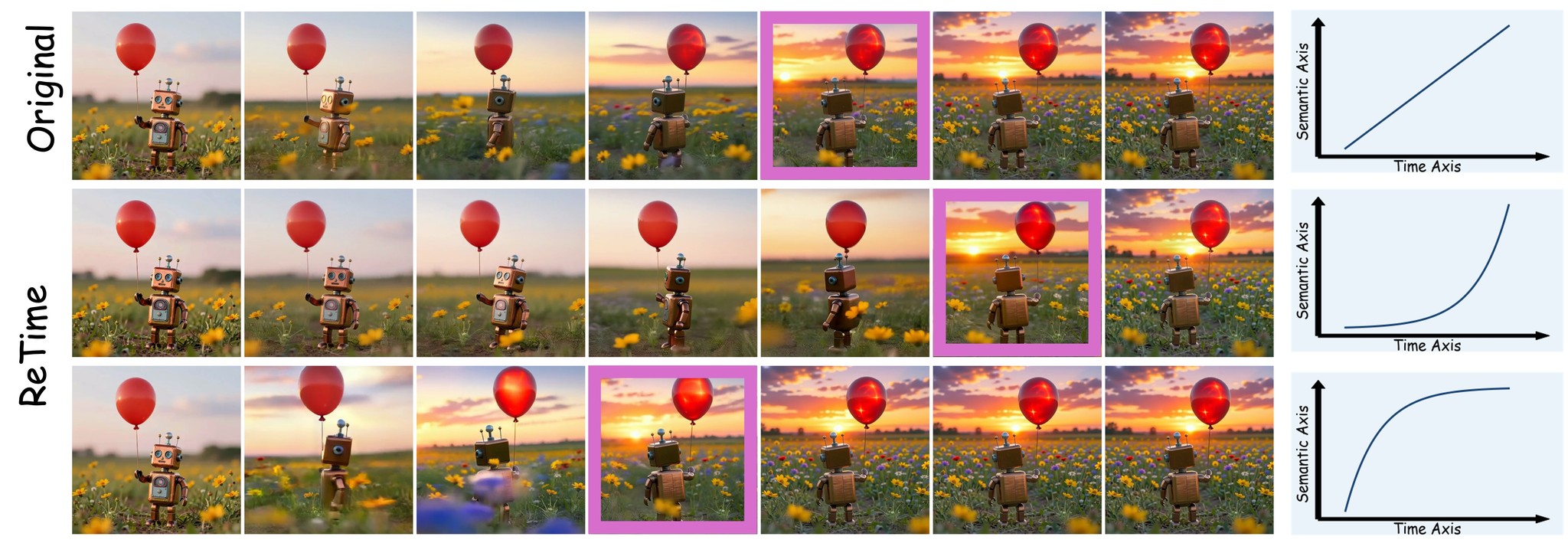}
  \caption{\textbf{Non-Linear Retiming.} Instead of linearizing the SPF, the video is retimed to match rising and falling exponential curves. The marked frames indicate the sun's entry, highlighting the acceleration and deceleration relative to the original video.}
  \label{fig:exp_up_exp_down}
\end{figure}

While our primary focus is linearizing the SPF to achieve constant semantic speed, our method also supports arbitrary target pacing functions.
\paragraph{Exponential Retiming.}
In Figure~\ref{fig:exp_up_exp_down}, we demonstrate this capability using rising and falling exponential functions. The moment the sun enters the frame serves as a visual cue, illustrating the acceleration and deceleration caused by our reparameterization.

\paragraph{Synthetic Validation.}
\label{subsec:rotating_spot}

\begin{figure}[t]
  \centering
  \includegraphics[width=0.8\linewidth]{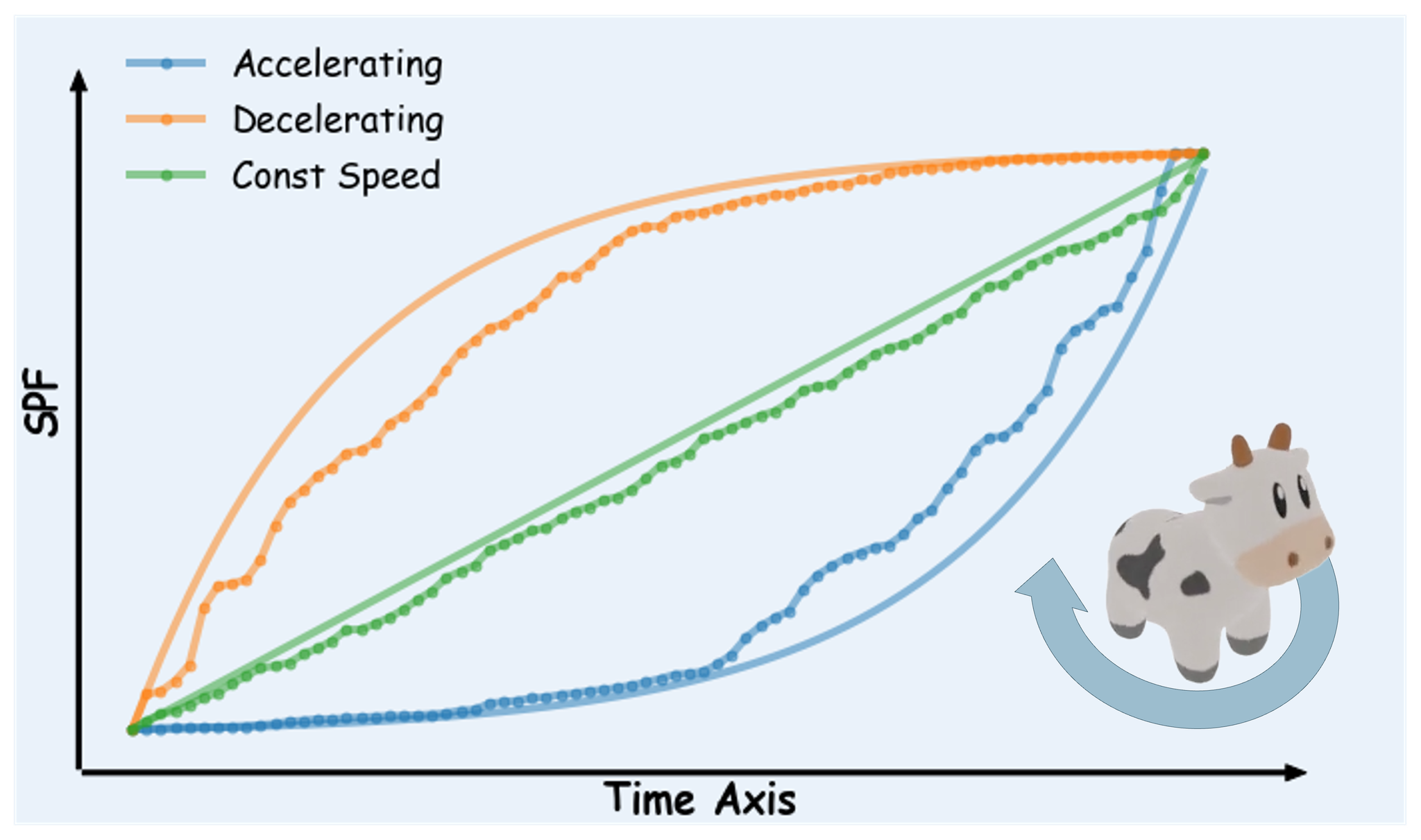}
  \caption{\textbf{Synthetic Validation.} Rotating-spot benchmark: angular position $\theta(t)$~(solid lines) and recovered SPF~(dotted lines) for constant, rising, and falling velocity profiles.}
  \label{fig:spot_potential}
\end{figure}

To validate that the SPF reflects cumulative semantic change, we generated synthetic videos of Keenan's Spot~\cite{crane2013robust} rotating on a white background under different angular velocity profiles: constant, rising exponential, and falling exponential.
This controlled experiment allows us to validate whether the SPF faithfully captures the pace induced by the different angular velocity profiles.
Furthermore, the minimalist setting of the synthetic scene, i.e., single object on a white background, ensures the SPF is affected mainly by the rotation pace.
As seen in Figure~\ref{fig:spot_potential}, the SPF (dotted lines) closely tracks the ground-truth angular position $\theta(t)$ (solid lines), mirroring the designed speed profiles. This confirms that the SPF captures the nonuniform pacing without relying on pixel-level motion.

\subsection{SPF Hyperparameter Ablation}

\begin{figure}[t]
  \centering
  \includegraphics[width=0.8\linewidth]{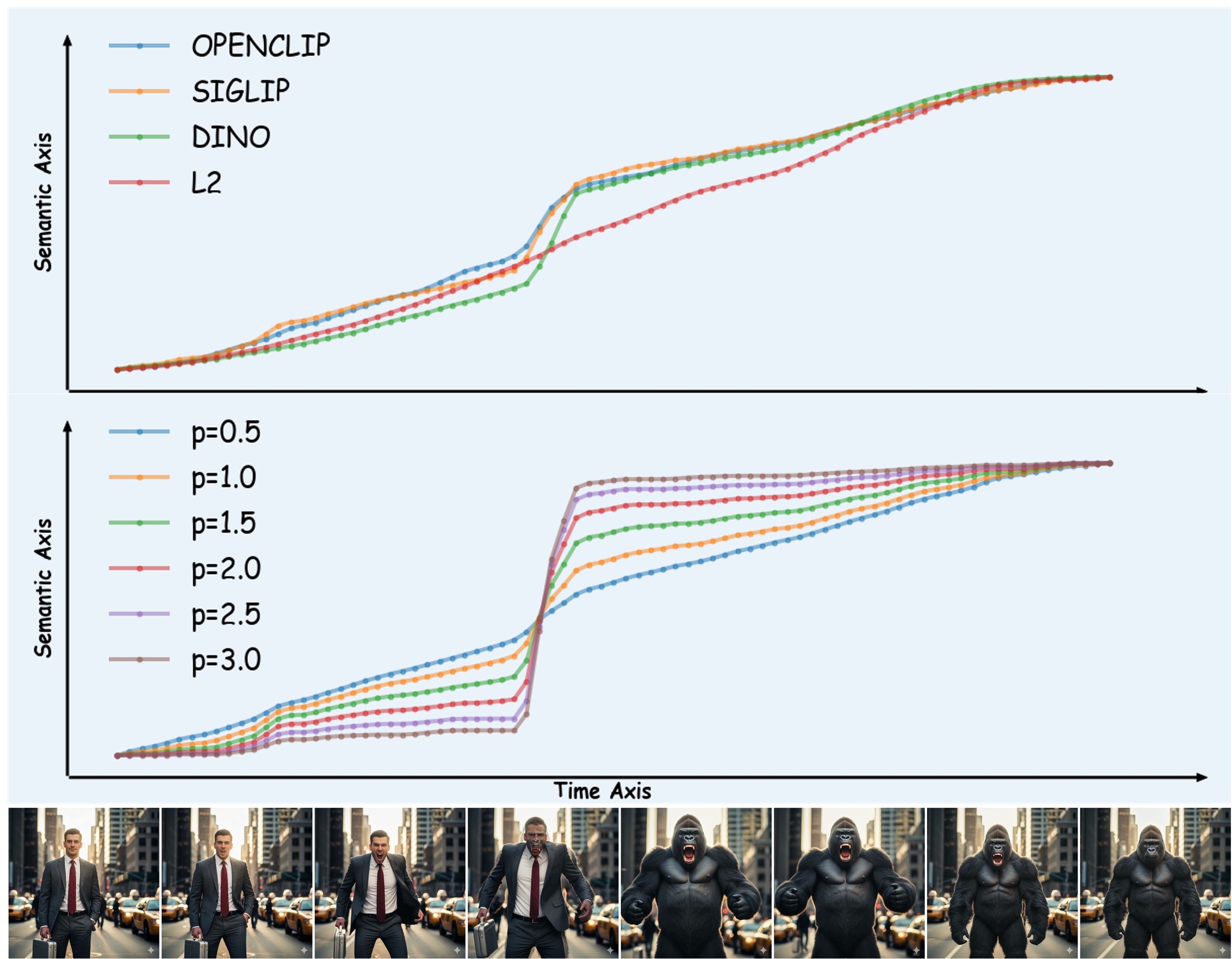}
  \caption{\textbf{SPF Ablation Study.} \textbf{Top:} Comparison of four pairwise models for computing the SPF. 
  The pixelwise L2 metric fails to capture the semantic shift, while SigLIP exhibits the best fine-grained sensitivity, detecting the onset of the man's anger. 
\textbf{Bottom:} Effect of the distance power $p$. Increasing $p$ acts as a contrast modulator for the semantic curve.}
  \label{fig:ablation}
\end{figure}

\paragraph{Pairwise Distance Model.} 
We investigate how the choice of frame embedding affects the SPF $S_i$. Specifically, we evaluate four representations: OpenCLIP (ViT-based contrastive), SigLIP (sigmoid loss contrastive), DINO (self-supervised), and a pixel-level baseline using $\ell_2$ distance. 
The results are presented in the top row of Figure~\ref{fig:ablation}. 
First, we observe that the $\ell_2$ metric fails to capture the rapid semantic transition of the man transforming into a gorilla, emphasizing the necessity of a semantic embedder, rather than relying on pixel-level differences.
While the remaining models yield comparable results, SigLIP demonstrates superior fine-grained sensitivity; notably, it exhibits a distinct local peak corresponding to the onset of the subject's anger, a nuance missed by the other metrics. 
Consequently, we adopt SigLIP as our default embedder as we empirically found it to produce the most perceptually aligned results.

\paragraph{Distance Power $p$.} 
We introduce the distance power hyperparameter $p$ to modulate the distance term via $\tilde{d}_{ij} = d_{ij}^p$ (Eq.~\eqref{eq:angular_distance}), analogous to a bilateral filter.
This parameter calibrates the semantic embeddings, which often preserve relative rank rather than absolute perceptual magnitude. As illustrated in the bottom of Figure~\ref{fig:ablation}, larger $p$ values increase the curve's contrast. While we typically default to $p=1$, we find that $p=2$ yields superior segmentation results for existing-video regeneration (Section~\ref{subsec:existing_video_linear}).
\subsection{Quantitative Evaluation}
\label{subsec:benchmark_options}
In the context of generated video retiming (Section~\ref{subsec:controlled_regen}), our approach is bounded by the generative capabilities of the base models (Wan2.2 or LTX-2). 
Because our method directly manipulates internal model embeddings, deviating from the models' standard inference protocols, it is imperative to verify that the visual fidelity of the output is maintained. 
We assess this using VBench~\cite{ji2024t2vbench} quality metrics, with quantitative results summarized in Table~\ref{tab:vbench_regen}.
\begin{table}[t]
\small
\setlength{\tabcolsep}{6pt}
\addtolength{\belowcaptionskip}{-6pt}
\centering
\caption{\textbf{Video Quality Preservation.}
  VBench evaluation on $N{=}128$ retimed videos per model. 
  Retiming results for both models maintain equivalent quality to the original input across all metrics.
}
\resizebox{\linewidth}{!}{%
\begin{tabular}{ll
                S[table-format=1.3]
                S[table-format=1.3]
                S[table-format=1.3]}
\toprule
& & \multicolumn{3}{c}{\textbf{VBench Metrics} $\uparrow$} \\
\cmidrule(lr){3-5}
\textbf{Model} & \textbf{Type} & \textbf{Aesthetic Q.} & \textbf{Motion S.} & \textbf{Temporal F.} \\
\midrule
\multirow{2}{*}{Wan2.2} 
& Original       & \num{0.630 \pm 0.093} & \num{0.987 \pm 0.010} & \num{0.978 \pm 0.019} \\
& Retimed & \num{0.626 \pm 0.090} & \num{0.987 \pm 0.010} & \num{0.978 \pm 0.019} \\
\midrule
\multirow{2}{*}{LTX-2} 
& Original    & \num{0.660 \pm 0.085} & \num{0.994 \pm 0.003} & \num{0.990 \pm 0.008} \\
& Retimed     & \num{0.656 \pm 0.087} & \num{0.993 \pm 0.005} & \num{0.990 \pm 0.009} \\
\bottomrule
\end{tabular}
}
\vspace{0.3cm}

\label{tab:vbench_regen}
\end{table}

Across all metrics, the retimed videos fall within one standard deviation of the baseline, indicating that our temporal manipulation preserves visual fidelity.
Additionally, we have conducted a subjective user study that confirms our method significantly improves semantic pacing (88\% preference) while maintaining visual quality. See the supplementary material for the full details.
\revision{To quantify pacing, we introduce an SPF-based \emph{Linearity Score} that measures how semantic progress follows an ideal linear pace (see supplementary, Section~E).}

\section{Conclusions, Limitations, and Future Work}

We introduced the \emph{Semantic Progress Function}, a simple and interpretable representation that captures how meaning evolves over time in image and video sequences. By reducing a complex transformation to a one-dimensional semantic trajectory, the proposed framework makes it possible to explicitly measure semantic pacing, identify abrupt transitions, and compare temporal behavior across different generative processes in a model-agnostic manner.

Building on this analysis, we proposed \emph{semantic linearization}, a principled remedy that reparameterizes time so that semantic change unfolds at a constant rate. We demonstrated two complementary realizations of this idea: direct intervention during generation via temporally warped positional embeddings, and post-hoc linearization of existing videos through segmented regeneration. Together, these techniques enable smoother and more predictable transformations without requiring model retraining or manual annotation.
\paragraph{Limitations.}
The proposed semantic analysis relies on frame-level embeddings, and as a result, may be influenced by rapid camera motion, strong lighting changes, or large non-semantic appearance variations that affect the embedding space. In such cases, the estimated progress function may partially reflect perceptual change rather than pure semantic evolution. While our local weighting formulation mitigates some of these effects, fully disentangling motion, appearance, and semantics remains an open challenge.
In addition, the iterative refinement introduced in Section~\ref{subsec:controlled_regen} progressively shifts temporal embeddings away from their trained distribution, which may degrade output quality if too many iterations are applied.
\paragraph{Future Work.}
Several directions emerge from this work. First, incorporating motion-aware or temporally grounded embeddings may improve robustness in highly dynamic scenes.
Second, extending the framework to jointly analyze multiple semantic dimensions, for example, disentangling semantic factors such as identity, style, and geometry, could enable richer control over different aspects of change.
Finally, semantic progress analysis may serve as a foundation for a range of other downstream applications, including benchmarking temporal behavior of generative models, keyframe-based video summarization, video semantic thumbnailing, and more.
\revision{Moreover, linearized morphs naturally produce uniformly paced semantic trajectories between two visual states, positioning our framework as a data generation tool for training \textit{edit strength} controlled models.}
We hope this perspective contributes to a deeper understanding and improved controllability of temporal behavior in generative models.

\begin{acks}
    We would like to thank Elad Richardson and Yuval Alaluf for their early feedback and insightful discussions. 
    This work was supported by the Center for AI and Data Science at Tel Aviv University (TAD) and the Israel Science Foundation under Grant No.~\grantnum{}{2492/20} and Grant No.~\grantnum{}{1473/24}.
\end{acks}

\bibliographystyle{ACM-Reference-Format}
\bibliography{bib}

\newpage

\begin{figure*}[t!] %
    \centering
    \includegraphics[width=\textwidth]{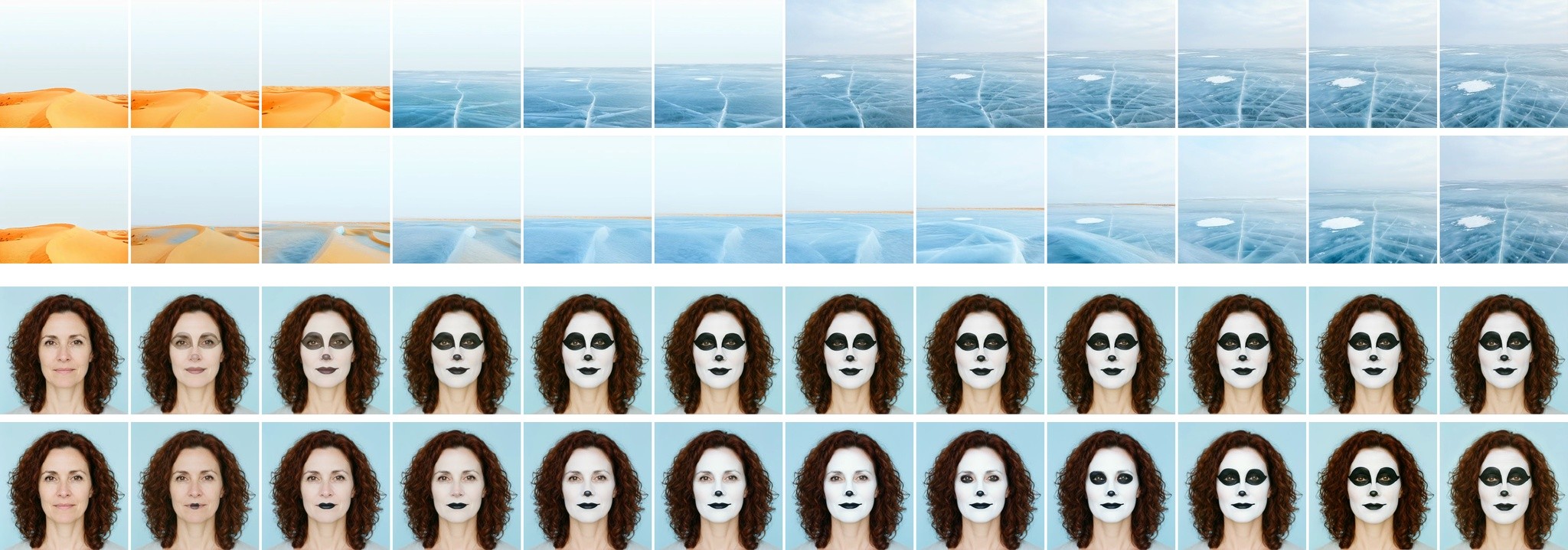}
    \caption{\textbf{Complex Scene Linearization.} Our method effectively handles diverse semantic scales, from global lighting shifts (\textbf{Top}: landscape) to fine-grained structural evolution (\textbf{Bottom}: human face), creating smooth progressions without artifacts.}
    \label{fig:general_results_1}
        
    \includegraphics[width=\textwidth]{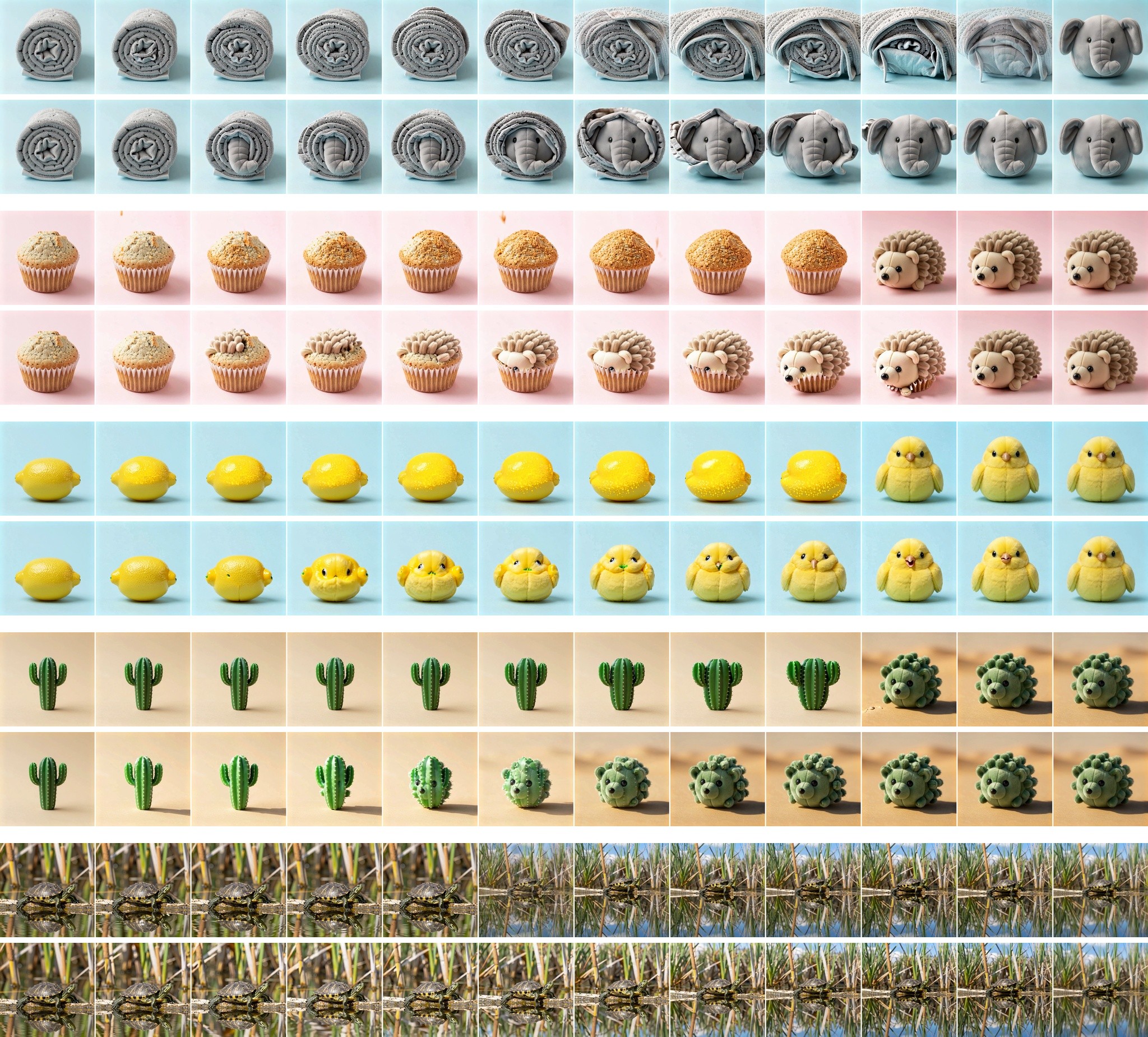}
    \caption{\textbf{Generalization to LTX-2.} Application of our ReTime framework to LTX-2~\shortcite{hacohen2025ltx2}. The successful linearization, despite architectural differences from Wan2.2, confirms the model-agnostic applicability of the Semantic Progress Function.}
    \label{fig:general_results_ltx}
\end{figure*}

\end{document}